\documentclass[sigconf]{acmart}
\renewcommand\footnotetextcopyrightpermission[1]{}
\setcopyright{none}

\AtBeginDocument{%
  }

\setcopyright{acmlicensed}
\copyrightyear{2018}
\acmYear{2018}
\acmDOI{XXXXXXX.XXXXXXX}
\acmConference[Conference acronym 'XX]{Make sure to enter the correct
  conference title from your rights confirmation email}{June 03--05,
  2018}{Woodstock, NY}
\acmISBN{978-1-4503-XXXX-X/2018/06}
  
\usepackage{subcaption}
\usepackage{booktabs}
\usepackage{multirow}
\usepackage{array}
\usepackage{algorithm}
\usepackage{algpseudocode}
\usepackage[most]{tcolorbox}
\usepackage{amsmath}
\usepackage{bm}
\usepackage[table,xcdraw]{xcolor}
\usepackage{booktabs}
\usepackage{multirow}
\usepackage{makecell}
\usepackage{xcolor}

\definecolor{mygreen}{RGB}{0,128,0}
\definecolor{myboxbg}{HTML}{F0EFFF}

\AtBeginDocument{
  }

\begin{document}

\title{ODYSSE: Episode-wise Policy Optimization for Personalized Agentic Reasoning}

\author{Jiaqi Zhang}
\affiliation{
  \institution{The University of Queensland}
  \city{Brisbane}
  \country{Australia}}
\email{jiaqi.zhang@uq.edu.au}

\author{Tong Chen}
\affiliation{
  \institution{The University of Queensland}
  \city{Brisbane}
  \country{Australia}}
\email{tong.chen@uq.edu.au}

\author{Junliang Yu}
\affiliation{
  \institution{Griffith University}
  \city{Gold Coast}
  \country{Australia}
  }
\email{junl.yu@outlook.com}

\author{Quoc Viet Hung Nguyen}
\affiliation{
  \institution{Griffith University}
  \city{Gold Coast}
  \country{Australia}}
\email{quocviethung.nguyen@griffith.edu.au}

\author{Hongzhi Yin}
\authornote{Corresponding author.}
\affiliation{
  \institution{The University of Queensland}
  \city{Brisbane}
  \country{Australia}}
\email{h.yin1@uq.edu.au}

\begin{abstract}
Agentic systems have rapidly advanced in their ability to interact with real-world environments, leverage external tools, and provide services for users. However, unlike natural-world tasks that assume well-defined instructions, human-centered scenarios are characterized by ambiguous requests that lead to large, open-ended solution spaces. Decoding users' personalized preferences is therefore essential for narrowing the candidate solution space. This introduces a new challenge, \textit{personalized agentic reasoning}, which requires agents to transform ambiguous user requests into concrete needs to deliver personalized services. In this paper, we present \textbf{ODYSSE}, a Reinforced Fine-Tuning (RFT) framework for personalized agentic reasoning. At its core, ODYSSE proposes \textbf{Episode-wise GRPO (ESPO)}, a novel extension of Group Relative Policy Optimization (GRPO) designed to address long action horizons and strong cross-step dependencies in personalized agentic reasoning. Rather than optimizing individual steps independently, ESPO introduces an episode-level reward mechanism together with episodic advantage estimation, enabling upstream evidence to effectively guide downstream personalized decisions and allowing agents to progressively resolve ambiguous user requests across multiple interaction steps. We further propose an episodic batch sampler that groups actions from the same episode into unified training batches, facilitating coherent optimization under ESPO. We evaluate ODYSSE on realistic long-horizon personalized GUI reasoning tasks. Experimental results demonstrate that ODYSSE consistently outperforms both specialist and general-purpose LVLMs, highlighting its effectiveness for personalized agentic reasoning.
\end{abstract}

\maketitle

\section{Introduction}
\label{sec:intro}
Agentic systems based on Large Vision-Language Models (LVLMs) have recently demonstrated strong capabilities in multimodal understanding, instruction following, and complex reasoning, making them increasingly promising for real-world decision-making tasks~\cite{wei2026agentic, li2024personal, li2026laboratory}. 
Powered by representative reasoning backbones such as 
GPT-o series~\cite{achiam2023gpt}, 
Google Gemini~\cite{comanici2025gemini}, and 
DeepSeek-R1~\cite{guo2025deepseek}, 
agentic systems are being deployed beyond simplified experimental environments into more realistic, adaptive, and human-centered service scenarios.
Such real-world applications span diverse domains, including 
digital personal assistants~\cite{zhang2026smartagent, wang2026openclaw, wang2026assistant}, 
web agents~\cite{wu2026webdancer, he2024webvoyager, chae2025web}, 
household service robots~\cite{Szot2021habitat, puig2024habitat, yenamandra2023homerobot}, etc. 

Across these scenarios, the role of agentic systems shifts from executing predefined instructions to inferring personalized goals for a particular user.
In many real-world services, users do not approach an agent with a fully determined task plan. Instead, they seek assistance to clarify available options, identify relevant trade-offs, and settle on their preferences.
This uncertainty often leads to an initial ambiguous request.
Decoding such requests places new demands on real-world agent design.
We refer to this capability as \textit{personalized agentic reasoning}, where agents transform an ambiguous user request into concrete personalized needs by 
narrowing the candidate solution space with user feedback.

However, existing agent designs are not aligned with such personalized requirements.
Most agents are designed to follow either clear instructions grounded in natural-world observations~\cite{luo2025gui, huang2024embodied}, or user requests pre-specified as direct commands~\cite{zhang2026smartagent, zhao2025appagent}.
This formulation works when the user's goal is explicit, but becomes insufficient when agents must handle ambiguous requests.
For example, if a user asks to locate “\texttt{Hilton New York Midtown at 1335 Avenue of the Americas}”, any existing GUI agent can do so by directly typing keywords and returning the unique result. 
Yet intuitively, such a request is far from how real users behave. 
A real request would simply be “\texttt{I am going to New York, any hotels?}”
Such an ambiguous request leaves an open-ended solution space, where the appropriate response varies with different users' personalized preferences.
As a result, existing agents lack effective mechanisms to perceive the personalized factor in ambiguous requests.

Another limitation arises from the optimization paradigm.
Real-world reasoning rarely depends on a single observation or single-turn question answering~\cite{wang2026rlanything}. 
Instead, it requires agents to integrate signals across long-horizon interactions, where personalized needs often unfold through longer interaction steps and stronger cross-step dependencies~\cite{zhou2024language, zhang2024ninerec}. 
However, existing agents are not optimized towards such long-horizon reasoning.
Firstly, many agents rely on outcome-based supervision, often using external functional rewards~\cite{dong2026agentic, zhou2025sweet, da2025agent}. This leaves optimization dominated by final-answer correctness, while the intermediate steps are underconstrained~\cite{lightman2024let, setlur2025rewarding}. Agents may forgo crucial intermediates or exploit shortcuts.
Second, many agents adopt step-wise optimization, which assumes that each interaction step can be optimized independently~\cite{deng2024novice, chen2023fireact, sinha2026the}.
This fragments cross-step dependencies, leaving agents struggle to connect upstream evidence with downstream personalized decisions~\cite{wang2026reasoning, kim2025reflact, sinha2026the}.
Overall, existing agents are ill-suited for coherent long-horizon reasoning.

In light of the deficiencies of existing solutions to personalized agentic reasoning, we present \textbf{ODYSSE}, which systematically reasons its way out of an initially brief and ambiguous user input to perform personalized actions.
To bypass the coarse feedback signals in existing solutions, 
instead of assigning supervision only to final outcomes or isolated actions, we propose \textit{episode-wise relative policy optimization} (ESPO), a novel RFT framework that allows ODYSSE to identify which exact steps are driving performance or causing error when eliciting personalized needs.
Specifically, after warm-up training with SFT, ESPO organizes an episode-aware reward mechanism in three components.
First, it introduces stage-specific verifiable rewards for individual action steps. These step-wise rewards provide reliable feedback for heterogeneous stages such as GUI execution, intent prediction, and personalized decision-making.
Second, ESPO constructs a Chain-of-User-Thought (COUT) reward from two complementary perspectives on how the user's ambiguous request is decoded: 
\textit{Intent Confidence} evaluates the confidence in the user intent inferred from preceding GUI actions,
while 
\textit{Intent Contribution} evaluates the contribution of the inferred intent to the final personalized decision. 
Built upon this bidirectional dependency, the COUT reward enables traceable decoding of ambiguous user requests.
Third, an episodic advantage estimation mechanism broadcasts each episode's COUT advantage to its action steps and combines it with their step-wise advantages to form episode-wise advantages, 
thereby enabling cross-step intent alignment.
Finally, the resulting episode-wise advantages are then used to update the policy.
To support coherent ESPO, we further design an episodic batch sampler. During training, the sampler treats each episode as the minimum indivisible sampling unit and groups samples from the same episode into a unified training batch. This ensures that dependent action steps within the same episode are optimized together, allowing ESPO to compute COUT rewards and perform policy updates over the full interaction trajectory rather than isolated action steps.

Our contributions are summarized as follows:
\vspace{-2pt}

\begin{itemize}
\item We formulate personalized agentic reasoning, where agent systems must handle ambiguous requests from real-world users. We identify key limitations of existing agents, namely limited ability to decode ambiguous requests and fragmented outcome-/step-wise optimization, and analyze how they become particularly severe in personalized scenarios.

\item We put forward ODYSSE, a dedicated framework for personalized agentic reasoning. In ODYSSE, we propose a novel RFT method, namely ESPO that offers fine-grained, per-step feedback. With ESPO, ODYSSE decodes ambiguous user requests by executing basic GUI actions, inferring GUI-related intent, and generating personalized recommendations.

\item Experiments show that ODYSSE achieves superior performance compared with both specialist GUI agents and general-purpose LVLMs, demonstrating the effectiveness and broader potential of episode-wise optimization for long-horizon personalized agentic reasoning.
\end{itemize}

\section{Related Work}
\label{sec:related_work}

\subsection{Real-World Agentic Reasoning}
Real-world agentic reasoning focuses on interacting with practical environments such as GUIs, Web Agents, and Coding Agents, rather than static image-text or video-text reasoning tasks~\cite{wei2026agentic, li2024personal, li2026laboratory, luo2025gui, zhang2025towards, zhang2025survey, yang2024swe}.
SeeClick~\cite{cheng2024seeclick} enables LVLM agents to identify and execute interface actions directly from screenshots and textual instructions, demonstrating the potential of multimodal LVLMs for real-world GUI control across various environments.
GUI-R1~\cite{luo2025gui} enables GUI agents with rule-based RL, improving long-horizon interaction planning and action reasoning in dynamic interface environments.
SmartAgent~\cite{zhang2026smartagent} presents a personal assistant agent that integrates GUI interaction and personalized recommendations, highlighting the importance of long-horizon personalization decision-making in real-world agentic systems.
\begin{figure*}[t]
  \centering
  \includegraphics[
    width=\textwidth,
    trim=0.0cm 0.0cm 3.6cm 12.6cm,
    clip
  ]{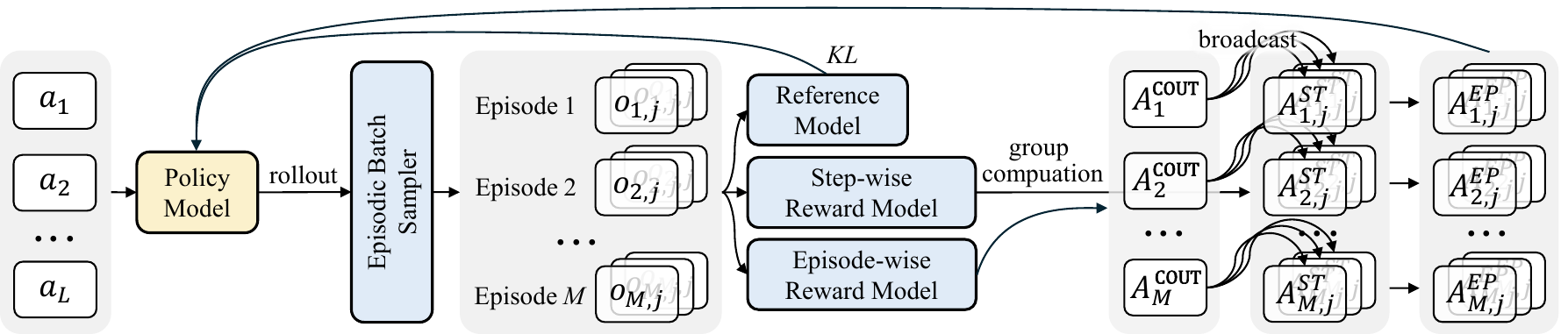}
  \caption{ESPO organizes $L$ isolated action steps into episodes. The stacked $o_{i,j}$ and $A_{i,j}$ blocks represent $G$ rollout-level quantities.}
\label{fig:Demonstration_ESPO}
\end{figure*}
Recent OpenClaw\footnote{\url{https://openclaw.ai/}} introduces an open-source agent framework that can serve as a virtual digital assistant on personal hardware.
The variants of OpenClaw further extend agentic systems toward real-time and personal adaptation in deployed environments~\cite{wang2026openclaw, ren2026x, zhu2026semaclaw}.
For example, OpenClaw-RL~\cite{wang2026openclaw} proposes a real-time personal agent framework built on a server-client architecture, enabling online agent learning by leveraging user re-queries, corrections, and explicit feedback.
Overall, existing real-world agents mainly extend agent-environment interaction, assuming simplified user instructions can be directly executed.
Our work instead studies \textit{personalized agentic reasoning}, where agents decode personalized needs from ambiguous requests through long-horizon interactions and generate personalized item recommendations.

\vspace{-6pt}

\subsection{Agentic Systems with Reinforcement Learning} 
Agentic systems learn to accomplish complex goals through autonomous environment perception and interaction. Unlike monolithic LLMs that typically perform CoT reasoning within a single forward generation process, agentic systems interact with evolving environments through iterative multi-turn actions, where each step corresponds to an environment-aware decision conditioned on updated observations~\cite{yao2022react, schick2023toolformer, wang2024survey}. 
Early agentic systems mainly relied on prompting engineering and SFT training, often suffer from compounding error accumulation, limited generalization, and weak long-term planning capabilities~\cite{chen2023fireact, yao2022react, schick2023toolformer}. 
Recent works~\cite{lambert2024tulu, luo2025gui, singh2025agentic} introduce Reinforcement Learning with Verifiable Rewards (RLVR). The RLVR setting improves interactive reasoning capabilities with deterministic reward functions, which are typically derived from executable environment feedback, such as task completion, correctness checking, or functional success signals. Existing methods have demonstrated that adapting RLVR to agentic systems through either step-wise rewards that assign verifiable supervision signals to intermediate interaction steps~\cite{yang2024swe, wang2026self}, or outcome-oriented rewards that only evaluate final task success~\cite{wang2026openclaw, lambert2024tulu}, can significantly improve agent reasoning and interaction capabilities in complex tasks.
However, outcome-oriented rewards only supervise final success, while step-wise rewards optimize actions separately. Our work instead augments step-wise rewards with episodic supervision to capture cross-step dependencies in personalized agentic reasoning.

\section{Problem Definition}
\label{sec:problem_definition}
In this work, we formulate \textit{Personalized Agentic Reasoning} in the context of digital personal assistants (e.g., flight booking and product purchasing assistants).
In this setting, an ambiguous request $q$ is resolved through agent-GUI (Graphical User Interface) interaction across heterogeneous action stages, including GUI execution, GUI-related user intent prediction, and personalized recommendation decision.
The agent learns from user demonstrations of this process, in which humans interact with an environment by operating the GUI to identify target items $\tau^\ast$ from its open-ended item pool $\mathcal{T}$.
Such a demonstration is represented as an interaction episode $e_i = \{a_{1}, a_{2}, \ldots, a_{N}\}$, where each action step $a_j$ is generated based on the current GUI observation $O_j\in\mathcal{O}$ and interaction history $H_j\in\mathcal{H}$.
These demonstrations teach the agent how interaction evidence guides personalized decisions.
Given a new ambiguous task, the agent is expected to: (1) generate appropriate actions $a_j \in e_i$ to interact with the environment, and (2) search, retrieve, and recommend candidate items $\tau \in \mathcal{T}$ that best satisfy the user's intent.
\section{Methodology}
\label{sec:methodology}

\subsection{Preliminaries}
Let $\mathcal{D} = \{e_{1}, e_{2}, \ldots, e_{M}\}$ denote the full episode dataset.
Given an action step $a_j \in e_i$, the corresponding training sample is denoted as $(q, O_j, H_j, y^\ast_j)$, with $y^\ast_j \in Y$ representing the corresponding ground-truth action.
Existing works optimize the agent either through SFT, which maximizes the likelihood of the ground-truth action:
\begin{equation}
\mathcal{L}_{\text{SFT}}
=
-\mathbb{E}_{(q,O_j,H_j,y_j)\sim\mathcal{D}}
\left[
\log \pi_{\theta}(y^\ast_j \mid q, O_j, H_j)
\right],
\label{eq:sft_objective}
\end{equation}
or through reinforcement learning, typically instantiated with GRPO~\cite{shao2024deepseekmath}.
For each sample, GRPO draws a group of candidate actions $\{y_1, y_2, \dots, y_G\}$ from the old policy $\pi_{\theta_{\mathrm{old}}}$ and estimates their advantages using relative rewards. 
The GRPO objective is formulated as:
\begin{equation}
\small
\begin{aligned}
\mathcal{J}_{\text{GRPO}}(\theta)=&
\mathbb{E}_{x\sim D,\{y_i\}_{i=1}^G\sim\pi_{\theta_{\text{old}}}}
\Bigg[
\frac{1}{G}\sum_{i=1}^G\frac{1}{|y_i|}\sum_{t=1}^{|y_i|}
\Bigg\{
\\
&\min\Bigg(
\frac{\pi_\theta(y_{i,t}|x,y_{i,<t})}
{\pi_{\theta_{\text{old}}}(y_{i,t}|x,y_{i,<t})} A_{i,t},
\\
&\text{clip}\Bigg(
\frac{\pi_\theta(y_{i,t}|x,y_{i,<t})}
{\pi_{\theta_{\text{old}}}(y_{i,t}|x,y_{i,<t})},
1-\epsilon,1+\epsilon
\Bigg) A_{i,t}
\Bigg)
\\
&-\beta \mathbb{D}_{\mathrm{KL}}
\left[\pi_\theta \| \pi_{\mathrm{ref}}\right]
\Bigg\}
\Bigg],
\end{aligned}
\label{eq:grpo}
\end{equation}
where $\beta$ balances reward maximization and KL regularization.
The advantage term \( \hat{A}_j \) is derived from the step-wise rewards of the group of responses \( \{R_1, R_2, \dots, R_l\} \) and calculated as:
\begin{equation}
A_i = \frac{r_i-\mathrm{mean}(\{r_1,r_2,\ldots,r_G\})}{\mathrm{std}(\{r_1,r_2,\ldots,r_G\})}.
\label{eq:advantages}
\end{equation}

\subsection{Episode-wise Policy Optimization (ESPO)} 
\label{sec:ESPO}
Existing agents are not explicitly optimized for such personalized reasoning processes, and the step-wise paradigm further makes it difficult to accumulate interaction evidence across the full episode and use it to guide downstream personalized decisions.
To bridge this gap, we propose \textbf{Episode-wise Policy Optimization (ESPO)}, as illustrated in Figure~\ref{fig:Demonstration_ESPO}. 
Specifically, ESPO preserves GRPO's group-relative update and augments step-wise supervision with episodic signals that explicitly model the dependencies among different action steps.
The overall training procedure is summarized in Algorithm~\ref{alg:episode_grpo}, and comprises of three main phases: ESPO first derives stage-specific verifiable rewards from heterogeneous action types (lines 4-9), then aggregates them into a Chain-of-User-Thought (COUT) reward to decode ambiguous user requests (lines 10-12), and finally broadcasts the resulting episodic supervision through an episode-wise advantage estimation mechanism (lines 13-17). In what follows, we unfold the design details of ESPO. 

\subsubsection{\textbf{Verifiable Rewards in Episodic Action Space}}
\label{sec:ESPO_reward}
Different steps within an episode serve distinct roles in personalized reasoning.
We therefore categorize action steps in each episode into three stages, including \textit{GUI task execution}, \textit{GUI-related user intent prediction}, and \textit{personalized decision}.
Each stage is associated with a specific action space and a corresponding verifiable reward.
This enables fine-grained supervision beyond final outcomes.

For all action steps $a_j \in e_i$, we define a format reward $R_{\mathrm{format}}$ to verify whether the generated response follows the required format:
\begin{equation}
R_\mathrm{{format}} =
\left\{
\begin{array}{ll}
0.5 & \text{if the format is correct}, \\
0.0 & \text{otherwise}.
\end{array}
\right.
\label{eq:R_format}
\end{equation}

For \textit{GUI task execution} steps, we define $R_{\mathrm{GUI\text{-}accuracy}}$ in Eq.~(\ref{eq:R_gui_accuracy}) as a fine-grained reward over the \texttt{<action type>} and \texttt{<action value>} parsed from the generated response.
\begin{equation}
R_{\mathrm{GUI\text{-}accuracy}} =
\left\{
\begin{array}{ll}
\;\;0.5 & \text{if both are correct}, \\
\;\;0.2 & \text{if only \texttt{<action type>} is correct}, \\
-0.2 & \text{if only \texttt{<action value>} is correct}, \\
-0.5 & \text{otherwise}.
\end{array}
\right.
\label{eq:R_gui_accuracy}
\end{equation}

\begin{algorithm}[t]
\small
\caption{Episode-wise GRPO (ESPO)}
\label{alg:episode_grpo}
\begin{algorithmic}[1]
\Statex \hspace{-\algorithmicindent}\textbf{Input:} Training dataset $\mathcal{D}$ with episode ids.
\Statex \hspace{-\algorithmicindent}\textbf{Parameter:} Policy $\pi_{\theta}$, training iterations $T$, KL coefficient $\beta$.
\Statex \hspace{-\algorithmicindent}\textbf{Output:} Optimized policy $\pi_{\theta}$.
\For{$t = 1$ to $T$}
    \State Sample a mini-batch $\mathcal{B} \in \mathcal{D}$ with complete episodes.
    \ForAll{episode $e_i \in \mathcal{B}$}   
        \ForAll{step $a_j$}
            \State Compute format reward $R_{\mathrm{format}}$;
            \State Compute accuracy reward $R_{\mathrm{accuracy}}$;
            \State Compute step-wise reward $R_{\mathrm{step\text{-}wise}}^{(j)}$ by Eq.~(\ref{eq:step_wise_reward});
            \State Compute intent matching score $Intent_{\mathrm{acc}}$ by Eq.~(\ref{eq:R_IntentAcc});
        \EndFor
        \State Aggregate GUI action gate $\phi$ by Eq.~(\ref{eq:R_phi_gate});
        \State Aggregate decision gate $\psi$ by Eq.~(\ref{eq:R_psi_gate});
        \State Compute $R_{\mathrm{COUT}}$ by Eq.~(\ref{eq:Rcout});
        \ForAll{step $a_j$}
            \State Compute step-wise advantage $A_{\mathrm{step\text{-}wise}}(a_j)$ by Eq.~(\ref{eq:step_wise_advantage});
            \State Broadcast $R_{\mathrm{COUT}}$ to step $a_j$ to obtain $A_{\mathrm{COUT}}(e_i)$ by Eq.~(\ref{eq:cout_advantage});
        \EndFor
        \State Compute episode-wise advantages ${A_{\mathrm{episode\text{-}wise}}^{(i,j)}}(e_i)$ by Eq.~(\ref{eq:episode_wise_advantage});
    \EndFor
    \State Update $\theta$ using GRPO with KL regularization;
\EndFor
\State \Return optimized policy $\pi_{\theta}$;
\end{algorithmic}
\end{algorithm}
For GUI-related user intent prediction steps, we define an intent matching reward $Intent_{\mathrm{acc}}$ in Eq.~(\ref{eq:R_IntentAcc}) to measure whether the predicted intent matches the user's ground-truth GUI intent $y^\ast_j$: 
\begin{equation}
Intent_{\mathrm{acc}} =
\left\{
\begin{array}{ll}
0.5 &
\begin{array}{@{}l@{}}
\text{if \texttt{<action type>} is correct and} \\
\mathrm{Cosine}
(\texttt{<action value>}, y^\ast_j) \geq \tau,
\end{array}
\\
0.0 & \text{otherwise}.
\end{array}
\right.
\label{eq:R_IntentAcc}
\end{equation}

For \textit{personalized decision} steps, we define $R_{\mathrm{dec\text{-}accuracy}}$ in Eq.~(\ref{eq:R_dec_accuracy}) to evaluate whether each item is correctly recommended.
This stage imposes a stricter criterion, requiring both \texttt{<action type>} and \texttt{<action value>} to be correct for a positive reward:
\begin{equation}
R_{\mathrm{dec\text{-}accuracy}} =
\left\{
\begin{array}{ll}
\;\;0.5 &
\begin{array}{@{}l@{}}
\text{if both \texttt{<action type>} and} \\
\text{\texttt{<action value>} are correct}
\end{array}\;,
\\
-0.5 & \text{otherwise}.
\end{array}
\right.
\label{eq:R_dec_accuracy}
\end{equation}

\subsubsection{\textbf{Chain-of-User-Thought (COUT) Reward}}
\label{sec:ESPO_Rcout}
Based on the stage-specific verifiable rewards, ESPO constructs a Chain-of-User-Thought reward $R_{\mathrm{COUT}}$ 
to evaluate whether an episode of actions coherently decodes an ambiguous request into a personalized need.
Accordingly, we aggregate stage-level evidence into two gates to model cross-step dependencies.
Specifically, the GUI-action gate $\phi$ summarizes the reliability of preceding GUI task execution actions for intent prediction, while the decision gate $\psi$ summarizes the extent to which the inferred intent contributes to the final personalized decision:
\begin{equation}
\phi_i = \sigma \left(
k \cdot
\frac{\sum_{j} R_{\mathrm{GUI\text{-}accuracy}}^{(j)}}
{R_{\mathrm{GUI\text{-}accuracy}}^{\max}}
\right),
\label{eq:R_phi_gate}
\end{equation}
\begin{equation}
\psi_i = \sigma \left(
k \cdot
\frac{\sum_{j} R_{\mathrm{dec\text{-}accuracy}}^{(j)}}
{R_{\mathrm{dec\text{-}accuracy}}^{\max}}
\right),
\label{eq:R_psi_gate}
\end{equation}
where $k$ is the temperature coefficient.
With these gates, $R_{\mathrm{COUT}}$ is formulated as two complementary components:
\begin{equation}
R_{\mathrm{COUT}}^{(i)}
=
\underbrace{\phi_i \cdot Intent_{\mathrm{acc}}}_{\text{Intent Confidence}}
+
\underbrace{\psi_i \cdot Intent_{\mathrm{acc}}}_{\text{Intent Contribution}},
\label{eq:Rcout}
\end{equation}
where the two terms respectively correspond to \textit{Intent Confidence} and \textit{Intent Contribution}, as detailed below.

\textit{Intent Confidence.}
The term $\phi \cdot Intent_{\mathrm{acc}}$ measures whether intent prediction is supported by reliable GUI task execution.
Specifically, accurate intent prediction should receive a high reward only when the preceding GUI actions are trustworthy, as quantified by the GUI-action gate $\phi$ in Eq.~(\ref{eq:R_phi_gate}).

\textit{Intent Contribution.}
The term $\psi \cdot Intent_{\mathrm{acc}}$ measures whether the predicted intent contributes to the final personalized decision.
A correct intent prediction should receive a higher reward only when it further leads to a valid personalized decision, as quantified by the decision gate $\psi$ in Eq.~(\ref{eq:R_psi_gate}).

Together, these two components connect aggregated upstream GUI evidence with downstream personalized decision-making, encouraging the model to reason over the whole episode rather than optimizing each action independently.

\subsubsection{\textbf{Episodic Advantage Estimation}}
\label{sec:ESPO_advantage_estimation}
ESPO realizes episodic advantage estimation to assign the personalized credit encoded in $R_{\mathrm{COUT}}$ to each step. 
Specifically, for each step $a_j \in e_i$, we preserve the original GRPO-style relative optimization by maintaining a stage-specific step-wise rewards and step-wise advantage:
\begin{equation}
\resizebox{0.9\columnwidth}{!}{$
R_{\mathrm{step\text{-}wise}}^{(j)}
=
\left\{
\begin{array}{l}
R_{\mathrm{format}}^{(j)} + R_{\mathrm{GUI\text{-}accuracy}}^{(j)}, 
\, \text{if } a_j \text{ is GUI task step}, \\[1.5mm]
R_{\mathrm{format}}^{(j)} + Intent_{\mathrm{acc}}, 
\quad\; \text{if } a_j \text{ is user intent step}, \\[1.5mm]
R_{\mathrm{format}}^{(j)} + R_{\mathrm{dec\text{-}accuracy}}^{(j)}, 
\; \text{if } a_j \text{ is personalized step},
\end{array}
\right.
$}
\label{eq:step_wise_reward}
\end{equation}

\begin{equation}
A_{\mathrm{step\text{-}wise}}(a_{j,g})
=
\frac{
R_{\mathrm{step\text{-}wise}}^{(j,g)}
-
\mathrm{mean}\left(\{R_{\mathrm{step\text{-}wise}}^{(j,g)}\}_{g=1}^{G}\right)
}{
\mathrm{std}\left(\{R_{\mathrm{step\text{-}wise}}^{(j,g)}\}_{g=1}^{G}\right)
}.
\label{eq:step_wise_advantage}
\end{equation}

Then, we define the COUT advantage by broadcasting the $R_{\mathrm{COUT}}^{(i)}$ to every rollout of each step $a_j$:
\begin{equation}
A_{\mathrm{COUT}}(e_i)
=
\operatorname{Broadcast}_{a_j \in e_i,\; g=1,\ldots,G}
\left(R_{\mathrm{COUT}}^{(i)}\right).
\label{eq:cout_advantage}
\end{equation}

The final episodic advantage for the $g$-th rollout of step $a_j$ is then merged as:
\begin{equation}
A_{\mathrm{episode\text{-}wise}}^{(i,j,g)}
=
A_{\mathrm{step\text{-}wise}}(a_{j,g})
+
w \cdot A_{\mathrm{COUT}}(e_i),
\quad
g=1,\ldots,G,
\label{eq:episode_wise_advantage}
\end{equation}
where $w$ controls the influence of the COUT advantage.
In this estimation, $A_{\mathrm{step\text{-}wise}}(a_{j,g})$ provides local step-wise credit, while $A_{\mathrm{COUT}}(e_i)$ broadcasts episode-level personalized signals to all steps, forming a personalized GRPO-style advantage.

\subsubsection{\textbf{ESPO Objective Function}}
\label{sec:ESPO_objective}
After estimating the episodic advantage, ESPO updates the policy on episode-preserving mini-batches, where each step $a_j \in e_i$ is associated with $G$ rollouts ${y_{j,g}}_{g=1}^{G}$.
The overall objective of ESPO is formulated as:
\begin{equation}
\small
\begin{aligned}
\mathcal{J}_{\mathrm{ESPO}}(\theta)
=&\;
\mathbb{E}_{\substack{
e_i\sim\mathcal{D},\;
\{y_{j,g}\}_{g=1}^{G}
\sim\pi_{\theta_{\mathrm{old}}}
}}
\Bigg[
\frac{1}{|e_i|}
\sum_{a_j\in e_i}
\frac{1}{G}
\sum_{g=1}^{G}
\frac{1}{|y_{j,g}|}
\sum_{t=1}^{|y_{j,g}|}
\Bigg\{
\\
&\quad
\min\Bigg(
\rho_{j,g,t}(\theta)
A_{\mathrm{episode\text{-}wise}}^{(i,j,g)},
\\
&\quad
\mathrm{clip}\Big(
\rho_{j,g,t}(\theta),
1-\epsilon,
1+\epsilon
\Big)
A_{\mathrm{episode\text{-}wise}}^{(i,j,g)}
\Bigg)
\\
&\quad
-\beta\,
\mathbb{D}_{\mathrm{KL}}
\left[
\pi_\theta
\|
\pi_{\mathrm{ref}}
\right]
\Bigg\}
\Bigg],
\end{aligned}
\label{eq:espo_objective}
\end{equation}
where $\rho_{j,i,t}(\theta)$ is the importance ratio:
\begin{equation}
\begin{aligned}
\rho_{j,g,t}(\theta)
=
\frac{
\pi_\theta
\left(
y_{j,g,t}
\mid
x_j,
y_{j,g,<t}
\right)
}{
\pi_{\theta_{\mathrm{old}}}
\left(
y_{j,g,t}
\mid
x_j,
y_{j,g,<t}
\right)
},
\end{aligned}
\label{eq:importance_ratio}
\end{equation}
and, $\epsilon$ is the clipping parameter to
constrain policy updates,and $\beta$ is the coefficient for the KL-divergence penalty term $\mathbb{D}_{\mathrm{KL}}$.

Compared with standard GRPO-style optimization over isolated action steps, ESPO performs episode-preserving optimization with personalized episodic advantage estimation.
The former keeps steps from the same trajectory intact during policy updates, while the latter broadcasts episode-level personalized signals into the advantage. 
Therefore, ESPO encourages actions that are not only locally executable at the current GUI state but also beneficial to the overall personalized reasoning trajectory.

\subsection{Episode-wise Training Strategy} 
\label{sec:Training}

\subsubsection{\textbf{RFT Training}} 
\label{sec:Training_RFT}
ODYSSE follows a two-stage reinforced fine-tuning (RFT) pipeline. In the first stage, we perform SFT on expert action trajectories to cold-start the policy. 
This stage teaches the model to generate structured GUI actions conditioned on the user query, current observation, and execution history.
The SFT objective follows Eq.~(\ref{eq:sft_objective}), where each sample is an action-level tuple $(q,O_j,H_j,y_j^\ast)$ extracted from an episode.

In the second stage, we initialize the policy from the SFT checkpoint and optimize it with our ESPO. For each sampled episode, the model predicts actions conditioned on the current observation and action history, receives stage-specific verifiable rewards, and then obtains the episode-wise reward by Eq.~(\ref{eq:episode_wise_advantage}). These rewards are used to compute group-relative advantages and update the policy with the GRPO objective in Eq.~(\ref{eq:espo_objective}). Compared with directly applying GRPO to isolated steps, this RFT design preserves the stability of SFT while using episode-level feedback to improve long-horizon intent inference and personalized decision making.

\subsubsection{\textbf{Episodic Batch Sampler}}
\label{sec:Training_Sampler}
Standard batch construction~\cite{shao2024deepseekmath, schulman2017proximal, engstrom2020implementation} in GRPO-based training randomly shuffles individual action steps, which breaks the temporal structure required by episode-wise reward computation. To support ESPO, we design an in-batch episode sampler that treats each complete episode as the minimum sampling unit. 
Specifically, the sampler assigns each step an episode identifier indicating the episode to which it belongs.
Based on these identifiers, the sampler first groups all steps belonging to the same episode, preserves their original temporal order, and then shuffles episodes rather than individual steps.

The sampler subsequently constructs a training batch by sampling one or more complete episodes. Formally, the resulting batch can be represented as:
\begin{equation}
\mathcal B
=
\bigcup_{k=1}^{K} e_k ,
\quad
e_i \cap e_j = \varnothing,
\quad i\neq j,
\label{eq:episodic_batch_disjoint}
\end{equation}
where each sampled episode is represented as
\begin{equation}
e_k=\{a_1,a_2,\ldots,a_{N_k}\},
\label{eq:episode_structure}
\end{equation}
and each action step belongs to exactly one episode, i.e.,
\begin{equation}
a_j \notin e_m \cap e_n,
\qquad m\neq n.
\label{eq:no_cross_episode}
\end{equation}
where each episode remains intact throughout the sampling process. 
This preserves the integrity of each episode while preventing interference across different episodes, thereby enabling coherent episode-wise reward computation and credit assignment.

\begin{table}[t]
\centering
\small
\caption{Action spaces and output formats in SmartSpot.}
\vspace{-0.10in}
\label{tab:dataset_statistics}
\begin{tabular}{p{1.8cm}<{\centering}p{6.0cm}}
\toprule
\textbf{Stage} & \textbf{Expected Output} \\
\midrule
\multirow{3}{*}{Search} & <action type>: GUI operation (e.g., \texttt{\{click\}}, \texttt{\{type\}}); <action value>: (e.g., ``click point <0.23, 0.74>, type 'Next Page''') \\
\midrule
\multirow{3}{*}{Item Pool} &  <action type>: \texttt{\{pool\_found\}}; <action value>: Natural-language output (e.g., ``The user needs an economy class flight from London to New York'') \\
\midrule
\multirow{1}{*}{Recommend} & \makecell[l]{
<action type>: \texttt{\{recommendation\}};\\
<action value>: \texttt{\{Yes/No\}}
} \\
\bottomrule
\end{tabular}
\vskip -0.13in
\end{table}
\section{Experiments}
\label{sec:experiments}
In this section, we progressively evaluate ODYSSE by answering the following research questions:
\begin{itemize}
\item \textbf{RQ1}:Is RFT superior to SFT for personalized reasoning?
\item \textbf{RQ2}:If so, can ODYSSE outperform existing LVLMs? 
\item \textbf{RQ3}:Is ESPO responsible for ODYSSE's gains?
\item \textbf{RQ4}: What drives the effectiveness of ESPO?
\end{itemize}
Furthermore, we investigate the sensitivity of ODYSSE to different hyperparameter settings.
We also provide a case study in Appendix~\ref{app:Case_Study}.

\subsection{Experimental Setup}
\label{sec:experiment_setup}

\subsubsection{\textbf{Datasets.}}
We conduct experiments on SmartSpot~\cite{zhang2026smartagent}, a benchmark for long-horizon personalized reasoning. 
SmartSpot is organized into multiple episodes, each associated with an ambiguous user request expressing a real-world need, such as \texttt{I am going to New York, any hotels?} or \texttt{find me some choices about flights to Sydney}.
\begin{table*}[t]
\caption{Comparison with Generalist LVLMs and Specialist Agents on SmartSpot.}
\vspace{-0.10in}
\label{tab:main_results}
\centering
\begin{tabular}{p{3.6cm}<{\centering} p{3.2cm} p{2.2cm}<{\centering} p{2.2cm}<{\centering} p{2.2cm}<{\centering} p{2.2cm}<{\centering}}
\toprule
Category & Model & EleAcc & Op F1 & Step SR &RecAcc \\
\midrule
\multirow{11}{*}{Generalist LVLMs}
& Qwen3-VL-2B           & 0.03405 & 0.57179 & 0.02287 & 0.00000 \\
& Qwen3-VL-4B           & 0.02770 & 0.27256 & 0.01620 & 0.00000 \\
& Qwen3-VL-8B           & 0.04314 & 0.51566 & \underline{0.13800} & 0.23928 \\
\cmidrule{2-6}
& llava-1.5-7b          & 0.00000 & 0.24791 & 0.05749 & 0.14571 \\
& llava-v1.6-mistral-7b & 0.00000 & 0.32983 & 0.06911 & 0.23214 \\
& llava-v1.6-vicuna-7b  & 0.00000 & 0.22554 & 0.08537 & \underline{0.26642} \\
\cmidrule{2-6}
& InternVL3-1B          & 0.00000 & 0.35955 & 0.00000 & 0.00000 \\
& InternVL3-2B          & 0.03611 & 0.32854 & 0.06375 & 0.15285 \\
& InternVL3-8B          & 0.02727 & 0.53889 & 0.11851 & 0.24214 \\
\midrule
\multirow{3}{*}{Specialist Agents}
& SeeClick    & 0.08658 & 0.16162 & 0.09787 & 0.09500 \\
& SmartAgent  & 0.02202 & 0.39246 & 0.06675 & 0.00000 \\
& GUI-R1      & \underline{0.10637} & \underline{0.77512} & 0.13739 & 0.25500 \\
\midrule
Ours
& \textbf{ODYSSE} 
& \textbf{0.13229} {\textcolor{mygreen}{(+24.4\%)}} 
& \textbf{0.82405} {\textcolor{mygreen}{(+6.3\%)}} 
& \textbf{0.16119} {\textcolor{mygreen}{(+16.8\%)}} 
& \textbf{0.28571} {\textcolor{mygreen}{(+7.2\%)}} \\
\bottomrule
\end{tabular}
\end{table*}
An episode consists of a sequence of action steps, resulting in 1,713 steps across 102 episodes  (an average of 17 steps per episode).
Each episode spans three stages: \textit{Search} steps for navigating and retrieving an item pool, an \textit{Item Pool} step corresponding to a GUI page of the retrieved item list, and \textit{Recommend} steps for deciding whether each candidate should be recommended. 
Across all steps, the agent receives an ambiguous user request, a GUI screenshot, and previous actions as input.
The stage-specific action formats are summarized in Table~\ref{tab:dataset_statistics}.
We split the dataset by episode with an 8:1:1 ratio.
A representative data case is provided in Appendix~\ref{app:Case_Study}.

\subsubsection{\textbf{Evaluation Settings.}}
Following SmartAgent~\cite{zhang2026smartagent}, we evaluate model performance with two groups of metrics that progressively assess an agent's capabilities from low-level GUI interaction to high-level personalized reasoning.
\textit{Element Accuracy} (EleAcc), 
\textit{Operation F1 Score} (Op F1), and 
\textit{Step Success Rate} (Step SR) measure GUI actions from GUI element grounding accuracy, overall action matching, and step success, respectively.
\textit{Recommendation Accuracy} (RecAcc) evaluates whether the agent makes the correct final recommendation decision.

\begin{figure}[t]
  \centering
  \vspace{4pt}
  \includegraphics[width=\linewidth]{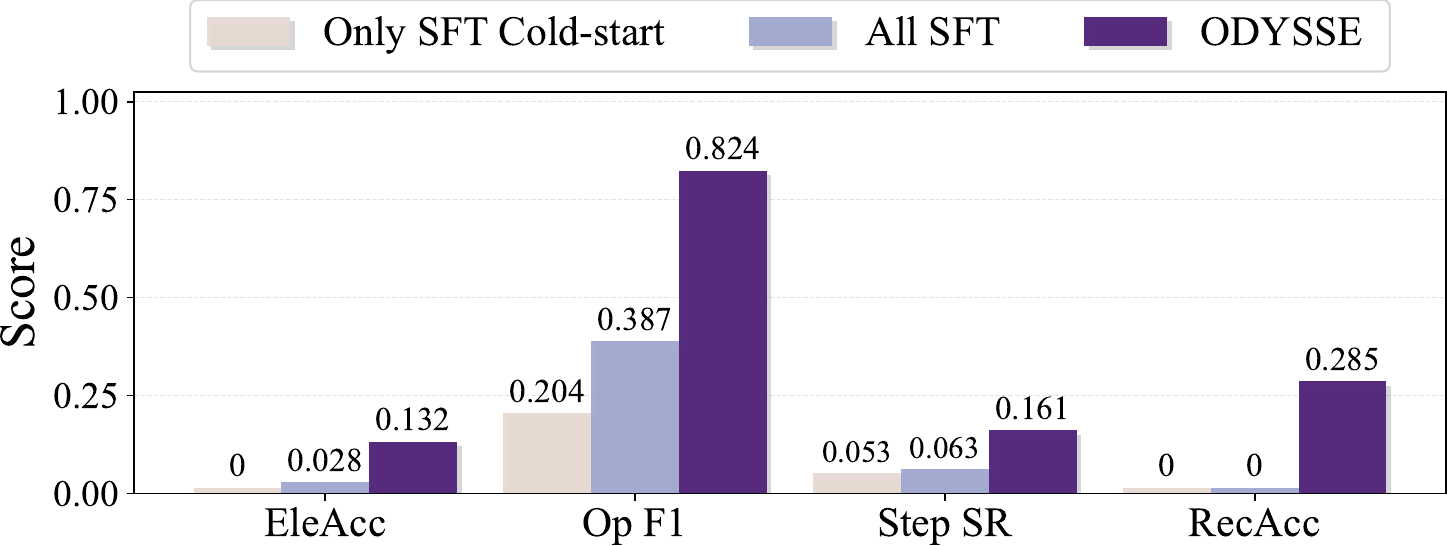}
  \vspace{-14pt}
  \caption{Our RFT-based ODYSSE vs. its SFT variants.}
  \label{fig:RFT_vs_SFT}
  \vspace{-14pt}
\end{figure}

\subsubsection{\textbf{Implementation Details.}}
\label{sec:implementation}
We use Qwen2.5-VL-3B as our base model. We first perform one epoch of SFT, followed by six epochs of ESPO implemented with EasyR1~\cite{zheng2025easyr1}. We enable in-batch sampling with one episode per batch and apply both the temperature $k$ in Eqs.~(\ref{eq:R_phi_gate}) and~(\ref{eq:R_psi_gate}) and advantage weight $w$ in Eq.~(\ref{eq:episode_wise_advantage}) to 1.0.
All experiments are conducted on a single NVIDIA H100-80G GPU.

\subsection{Preliminary Investigation (RQ1)}
\label{sec:RFT_vs_SFT}
We first study whether RFT offers a better training foundation than SFT for personalized agentic reasoning. 
SFT is widely used for learning expert behaviors, but its local imitation objective provides limited supervision for delayed feedback and episode-level outcomes. 
Given the recent success of RFT in reasoning and agentic tasks, we evaluate whether ODYSSE benefits from reinforcement feedback beyond purely SFT.

We examine this question with two SFT-based variants.  \textit{ODYSSE (Only SFT Cold-start)} keeps only the SFT cold-start, isolating whether RFT provides learning signals beyond SFT imitation.
\textit{ODYSSE (All SFT)} extends the SFT stage to the same iterations as ODYSSE but without any RL stage, testing whether RFT is more effective than simply scaling the SFT counterpart. 
Together, these comparisons assess both the necessity of RL and the paradigm-level advantage of RFT over SFT.

As shown in Figure~\ref{fig:RFT_vs_SFT}, the most pronounced gap appears between ODYSSE and \textit{ODYSSE (Only SFT Cold-start)}, which performs the worst among all variants, with both \textit{EleAcc} and \textit{RecAcc} being zero.
This indicates that the SFT cold-start can help the model acquire basic response formats, but it still struggles to ground GUI actions to the correct elements, let alone make personalized final recommendations
Compared with \textit{ODYSSE (Only SFT Cold-start)}, \textit{ODYSSE (All SFT)} obtains further improvements, suggesting that additional SFT iterations can strengthen low-level GUI interaction ability.
However, both SFT variants still fail to make any breakthrough on \textit{RecAcc}.
This indicates that high-level personalized perceiving remains beyond the capability of the SFT paradigm, even with additional SFT training iterations.
In contrast, ODYSSE achieves the best performance across all metrics.
The advantage is especially clear on \textit{RecAcc}, where the RFT stage enables the policy to acquire personalized decision-making that cannot be obtained from the SFT paradigm.
These results support our choice of building ODYSSE upon the RFT paradigm and motivate the following experiments to further examine how episode-wise optimization benefits personalized agentic reasoning.

\subsection{Overall Performance (RQ2)}
\label{sec:overall_performance}
After establishing the advantage of RFT over pure SFT, we next benchmark ODYSSE on the SmartSpot dataset against existing LVLMs to evaluate its overall capability.
To make this comparison comprehensive, we consider two complementary groups of baselines.
\textit{Generalist LVLMs} examine whether broad multimodal understanding alone is sufficient for personalized factor perception without task-specific adaptation.
\textit{Specialist Agents}examine whether optimization for specific environments alone is sufficient for capturing personalized factors.
Together, these two groups test whether personalized agentic reasoning can be achieved by either general-purpose multimodal capability or environment-specific agents, thereby highlighting the effectiveness of ODYSSE.
\begin{figure}[t]
  \centering
  \includegraphics[width=\linewidth]{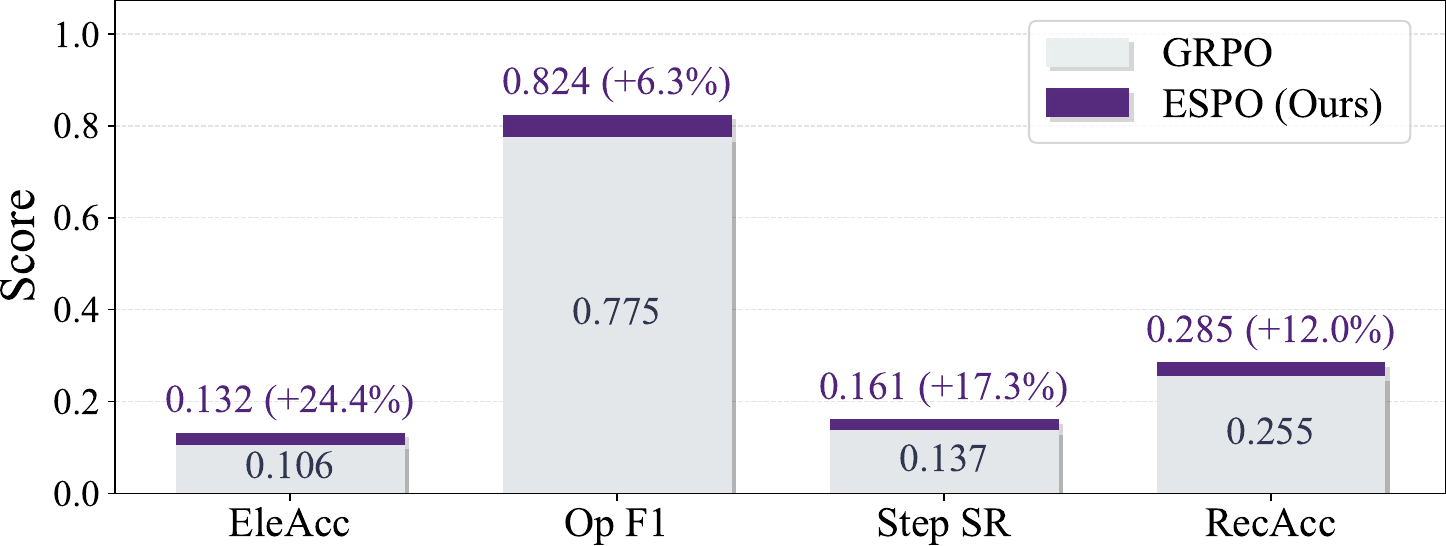}
  \vspace{-16pt}
  \caption{Our ESPO vs. GRPO.}
  \vspace{-10pt}
  \label{fig:ESPO_vs_GRPO}
\end{figure}
 
Specifically, the \textit{Generalist LVLMs} group includes:
\begin{itemize}

\item Qwen3-VL series~\cite{bai2025qwen3}, the latest Qwen model family, where we use 2B, 4B, and 8B variants to cover different scales.

\item LLaVA series~\cite{liu2023visual}, where we use three 7B variants with different language backbone.

\item InternVL3 series~\cite{zhu2025internvl3}, where we evaluate its 1B, 2B, and 8B variants.

\end{itemize}

\indent The \textit{Specialist Agents} group includes:
\begin{itemize}

\item SeeClick~\cite{cheng2024seeclick}, a representative GUI-specialized visual agent designed around SFT-based GUI. We reproduce it under the same fine-tuning epoch setting as ours.

\item SmartAgent~\cite{zhang2026smartagent}, a personalized agent that achieves state-of-the-art performance on the SmartSpot benchmark under the SFT paradigm.

\item GUI-R1~\cite{luo2025gui}, a recent leading GUI agent and the first RFT-based framework for GUI action prediction. We fine-tune its 3B version to maintain a comparable scale with ours.

\end{itemize}

As shown in Table~\ref{tab:main_results}, ODYSSE achieves the best performance across all metrics.
In particular, it improves the strongest baseline over $24\%$ on EleAcc and $16\%$ on Step SR, showing that ODYSSE can perform more accurate GUI element grounding and complete more successful steps.
Built upon these stronger low-level interaction abilities, ODYSSE further achieves a RecAcc of $28.57\%$, indicating its advantage in transferring GUI interaction evidence into final personalized recommendations.

\begin{figure}[t]
  \centering
  \includegraphics[width=\linewidth]{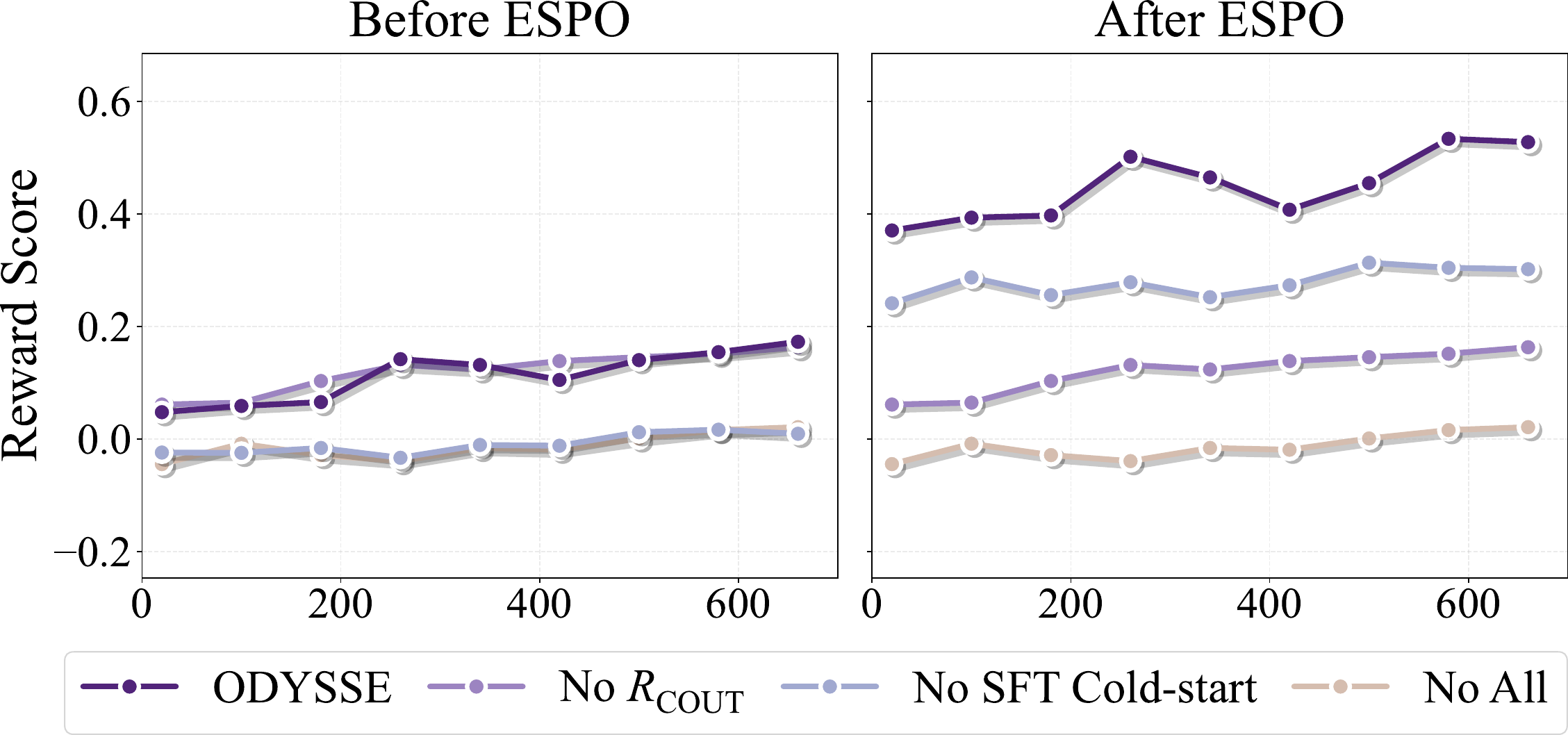}
  \caption{Ablation study for ODYSSE before (left) and after (right) our ESPO optimization. The y-axis represents the final reward score.}
  \label{fig:Ablation_Study}
  \vspace{-8pt}
\end{figure}

General-purpose LVLMs show clear limitations. 
Although some of them obtain non-zero RecAcc, their GUI interaction performance remains unconvincing.
For example, the LLaVA variants achieve relatively competitive RecAcc, but their EleAcc is consistently zero, suggesting that their final decisions are weakly grounded in accurate GUI element actions.
Similarly, Qwen3-VL-8B and InternVL3-8B show stronger performance than their smaller variants, but they still lag behind ODYSSE on both EleAcc and Step SR.

Specialist agents perform better but remain substantially behind ODYSSE.
The two SFT-based specialists, SeeClick and SmartAgent, show comparable performance on GUI action metrics, but both remain limited on RecAcc.
This suggests that SFT-based adaptation for GUI interaction can improve certain low-level abilities, but it is still insufficient for personalized decision-making.
The RFT-based GUI-R1 ranks second among nearly all baselines.
In contrast, equipped with ESPO, ODYSSE better guides the policy to decode personalized factors from interaction evidence, leading to the best performance across all metrics.

\subsection{In-Depth Analysis of ODYSSE (RQ3)}
\label{sec:in_depth_analysis}
Following the comparison with existing baselines, we further examine whether ODYSSE's gains are driven by its core optimization strategy, ESPO.
To isolate the effect of this design, we compare two controlled variants under the same backbone model, SFT cold-start, and reward setting: one trained with vanilla GRPO and the other with ESPO.

As shown in Figure~\ref{fig:ESPO_vs_GRPO}, replacing vanilla GRPO with ESPO brings consistent improvements across all metrics.
These improvements indicate that episode-wise optimization benefits both low-level GUI interaction and high-level personalized decision-making.
Specifically, the gains on EleAcc and Step SR suggest that ESPO helps the policy produce more reliable intermediate actions, while the improvement on RecAcc shows that these interaction gains are better translated into final personalized recommendations.

This comparison further explains why episode-wise optimization is necessary for long-horizon personalized agentic reasoning.
Vanilla GRPO still optimizes actions in a more local manner, providing limited guidance on how each step contributes to the complete episode.
In contrast, ESPO aligns intermediate GUI actions with episode-level outcomes, allowing the policy to better decode personalized interaction evidence across the trajectory.

Therefore, RQ3 further clarifies ESPO as the key source of ODYSSE's advantage.
Its improvement over vanilla GRPO confirms that episode-wise feedback is more effective for optimizing long-horizon personalized reasoning, which is consistent with the RFT advantage observed in RQ1 and further supports ODYSSE's strong overall performance in RQ2.

\subsection{Ablation Study (RQ4)}
\label{sec:ablation_study}
In this section, we conduct ablation studies to assess the contribution of each component and examine how ESPO reshapes their optimization dynamics.
Specifically, we organize the analysis into two levels: a framework-level ablation that evaluates the necessity of the major training components, and an internal-level ablation that further examines how detailed reward design, Intent Confidence $\phi$ and Intent Contribution $\psi$, inside $R_{\mathrm{COUT}}$ contribute to ESPO.

\begin{figure}[t]
  \centering
  \includegraphics[width=\linewidth]{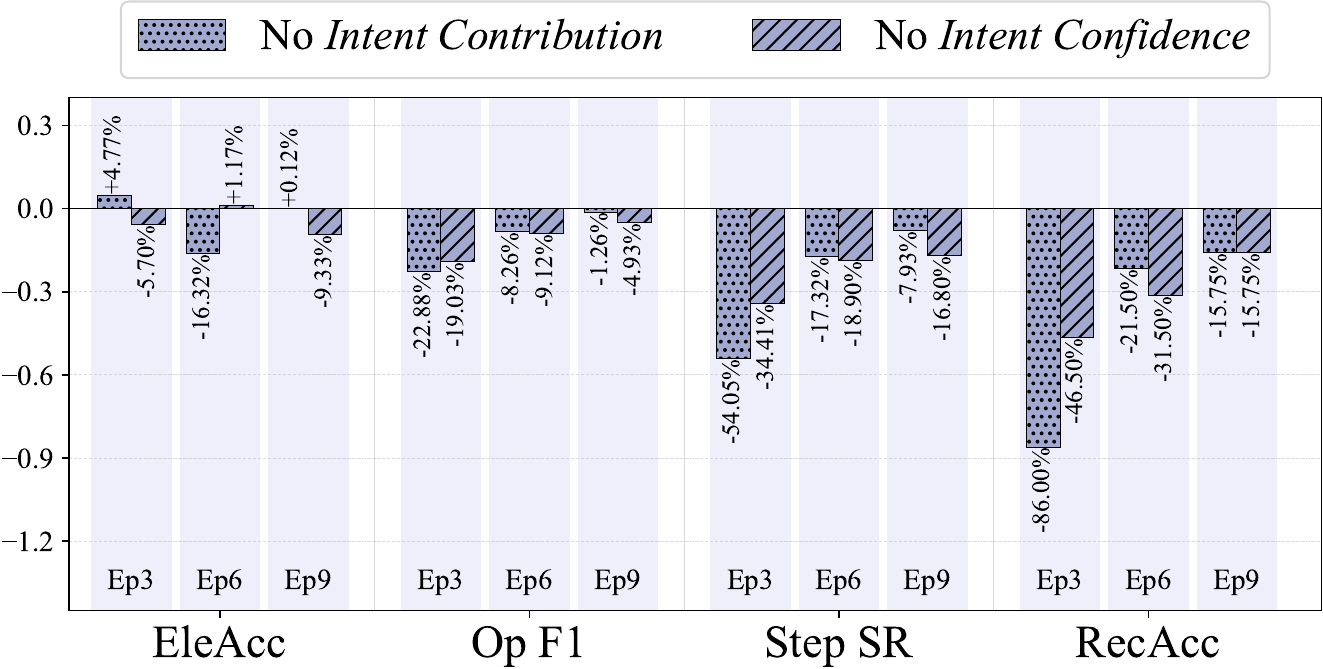}
  \vspace{-12pt}
  \caption{Ablation study of \textit{Intent Contribution} and  \textit{Intent Confidence} in $R_{\mathrm{COUT}}$. The y-axis presents the score $\Delta$ relative to ODYSSE (\%).}
  \label{fig:ablation_Phi_Psi}
  \vspace{-10pt}
\end{figure}

\begin{figure}[t]
  \centering
  \includegraphics[width=0.88\linewidth]{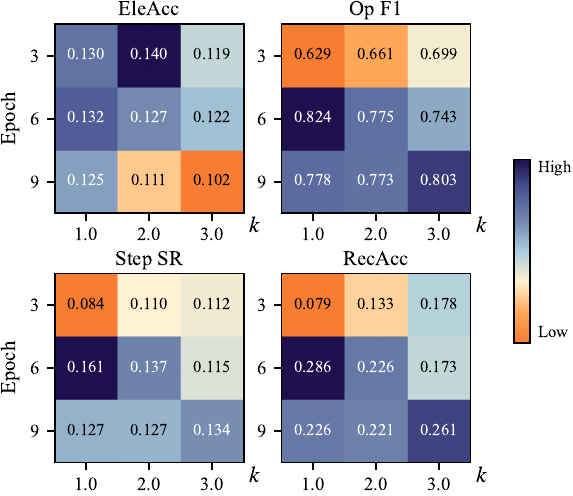}
  \vspace{-6pt}
  \caption{Hyper-parameter sensitivity analysis of ODYSSE under different Epoch and $k$ settings.}
  \label{fig:hyperparameter}
  \vspace{-10pt}
\end{figure}

\subsubsection{\textbf{Analysis of SFT Cold-start and $R_{\mathrm{COUT}}$}}
\label{sec:ablation_study_2}
At the framework level, we investigate the effects of SFT cold-start and the proposed $R_{\mathrm{COUT}}$ by constructing three variants:
We construct three ablation variants, including 
\textbf{\textit{No $\bm{R}_{\bm{\mathrm{COUT}}}$}},
\textbf{\textit{No SFT Cold-start}}, and
\textbf{\textit{No All}}, and compare them against the
\textbf{\textit{complete ODYSSE}} framework.
Figure~\ref{fig:Ablation_Study} presents their training dynamics by steps before and after applying ESPO. Several important observations can be drawn from the optimization dynamics:
\begin{itemize}
\item First, the four ablation variants are roughly separated into two groups before applying ESPO, where the presence or absence of SFT cold-start forms the dominant performance gap. In contrast, after introducing ESPO, the reward trajectories of different ablation variants become clearly separated, indicating that episodic policy optimization substantially improves the optimization discriminability among different architectural and reward designs.

\item More importantly, without ESPO, the reward trajectories of \textbf{\textit{complete ODYSSE}} and \textbf{\textit{No $\bm{R}_{\bm{\mathrm{COUT}}}$}} remain largely intertwined, with no clear performance gap between them. With ESPO, however, the \textbf{\textit{complete ODYSSE}} model becomes fully separated from \textbf{\textit{No $\bm{R}_{\bm{\mathrm{COUT}}}$}} and achieves a significantly larger reward advantage. This demonstrates that broadcasting the episode-wise reward provides effective long-horizon supervisory signals beyond conventional step-wise ones.

\item Furthermore, under ESPO optimization, \textbf{\textit{No SFT Cold-start}} noticeably surpasses \textbf{\textit{No $\bm{R}_{\bm{\mathrm{COUT}}}$}}, despite the absence of SFT cold-start. This suggests that episode-wise supervision introduced by ESPO can provide stronger long-horizon optimization benefits than relying solely on step-wise rewards with SFT cold-start.

\item Finally, the \textbf{\textit{No All}} variant consistently performs the worst after ESPO optimization. This indicates that the proposed components are not independently beneficial but instead work synergistically, jointly contributing to long-horizon personalized reasoning through complementary optimization effects.
\end{itemize}

\subsubsection{\textbf{Analysis of Intent Confidence and Intent Contribution}}
\label{sec:ablation_study_2}
Since ESPO's effectiveness mainly arises from $R_{\mathrm{COUT}}$, we further examine whether both components of $R_{\mathrm{COUT}}$, \textit{Intent Confidence} and \textit{Intent Contribution}, are necessary.
Specifically, we construct two variants by separately removing the \textit{Intent Confidence} term $\phi \cdot Intent_{\mathrm{acc}}$ and the \textit{Intent Contribution} term $\psi \cdot Intent_{\mathrm{acc}}$, respectively.

As shown in Figure~\ref{fig:ablation_Phi_Psi}, removing either term causes a cascading deterioration through the episode.
Although the effect on EleAcc is relatively mild, the gap grows when moving to Op F1 and Step SR, ultimately causing at least a $15\%$ gap on RecAcc and even an $86\%$ collapse at epoch 3.
This suggests that the bidirectional supervision slips toward a breakdown over the full episode.
Consequently, the variants gradually lose control over early GUI actions and later recommendation actions, which tend to optimize toward their own local objectives rather than mutually constraining each other across the episode.

Together, these results show that \textit{Intent Confidence} and \textit{Intent Contribution} are both indispensable for preserving the cross-stage dependency between GUI interaction and final personalized recommendation in $R_{\mathrm{COUT}}$.

\subsection{Hyperparameter Sensitivity Analysis}
\label{sec:Hyperparameter}
We further analyze the sensitivity of ODYSSE to two key hyperparameters: the training epoch and the temperature coefficient $k$ in $R_{\mathrm{COUT}}$.
In ESPO, $k$ controls the sharpness of the sigmoid gates for GUI-stage and recommendation-stage rewards.
A larger $k$ makes the gates more sensitive to accumulated rewards, thereby increasing the influence of $R_{\mathrm{COUT}}$ on episode-wise optimization.

As shown in Figure~\ref{fig:hyperparameter}, training ODYSSE for 6 epochs with $k=1.0$ yields leading results across all metrics.
This indicates that a smoother gate provides a more stable $R_{\mathrm{COUT}}$ for long-horizon optimization.
Larger $k$ may make the ESPO process more sensitive to partial or negative steps.
Some settings obtain isolated gains, such as higher \textit{EleAcc} at epoch 3 with $k=2.0$, but these improvements are limited to a single local metric and do not bring consistent benefits.
Similarly, extending training from 6 to 9 epochs does not further improve overall performance and even degrades several settings, suggesting possible over-optimization toward specific reward signals.
\section{Conclusion}
\label{sec:conclusion}
In this paper, we propose ODYSSE, a novel RFT framework for personalized agentic reasoning. To address ambiguous user requests in real-world scenarios, ODYSSE learns to decode personalized needs across long-horizon interactions. ODYSSE introduces ESPO to move beyond outcome-/step-wise optimization by enabling episode-wise learning over coherent interaction trajectories. Extensive experiments show that ODYSSE outperforms both specialist and general-purpose LVLMs, underscoring the importance of episode-wise optimization for personalized agentic reasoning.

\bibliographystyle{ACM-Reference-Format}
\bibliography{main}

\appendix

\newpage
\section{Appendix}
\label{sec:appendix}

\subsection{Detailed Prompt}
\label{app:prompt}

\begin{tcolorbox}[
    colback=myboxbg,
    colframe=black,
    arc=3mm,
    boxrule=1pt,
    enhanced,
    breakable,
    title=Prompt designs of our ODYSSE,
    coltitle=white,
    colbacktitle=black,
    fonttitle=\bfseries,
]
You are a personal GUI agent. In this UI screenshot \texttt{<image>}, I want you to continue executing the command \texttt{\{goal\}}, with the previous actions being \texttt{\{previous\_actions\}}. \\
Please provide exactly one next action. \\
The action must be one of: \\
\texttt{['click', 'type', 'scroll up', 'scroll down', 'scroll left', 'scroll right', 'exit', 'complete', 'recommendation']}. \\
Output the thinking process in \texttt{<think>} \texttt{</think>} tags, and the final answer in \texttt{<answer>} \texttt{</answer>} tags. \\
The final answer must contain exactly one dictionary in the following format: \\
\texttt{<think> ... </think>} \\
\texttt{<answer>[{'action': ACTION, 'point': [x, y], 'input\_text': TEXT}]</answer>} \\

Rules for each action: \\
\textbf{1. For \texttt{'click'}}:
\begin{itemize}
    \item output the click point \texttt{[x, y]}
    \item \texttt{input\_text} must be \texttt{'no input text'}
\end{itemize}

\textbf{2. For \texttt{'type'}}:
\begin{itemize}
    \item output the target point \texttt{[x, y]}
    \item \texttt{input\_text} must be the exact text to type
\end{itemize}

\textbf{3. For \texttt{'scroll up'}, \texttt{'scroll down'}, \texttt{'scroll left'}, \texttt{'scroll right'}, \texttt{'exit'}}:
\begin{itemize}
    \item \texttt{point} must be \texttt{[-100, -100]}
    \item \texttt{input\_text} must be \texttt{'no input text'}
\end{itemize}

\textbf{4. For \texttt{'pool\_found'}}:
\begin{itemize}
    \item \texttt{point} must be \texttt{[-100, -100]}
    \item \texttt{input\_text} must be a short summary of the user's final intent
\end{itemize}

\textbf{5. For \texttt{'recommendation'}}:
\begin{itemize}
    \item \texttt{point} must be \texttt{[-100, -100]}
    \item \texttt{input\_text} must be exactly one English word: \texttt{'Yes'} or \texttt{'No'}
    \item do not output any explanation, summary, or extra text
\end{itemize}

\textbf{Important constraints:}
\begin{itemize}
    \item Output exactly one action.
    \item Do not output multiple \texttt{<think>}/\texttt{<answer>} blocks.
    \item Do not output multiple dictionaries.
    \item For \texttt{'recommendation'}, \texttt{input\_text} must be exactly \texttt{'Yes'} or exactly \texttt{'No'}.
    \item For \texttt{'complete'}, \texttt{input\_text} must summarize the user's GUI intent in natural language.
\end{itemize}
\end{tcolorbox}

\clearpage
\onecolumn

\subsection{Case Study}
\label{app:Case_Study}

\begin{figure}[h]
  \centering
  \includegraphics[
    width=0.9\textwidth,
    trim=0.0cm 1.7cm 0cm 0cm,
    clip
  ]{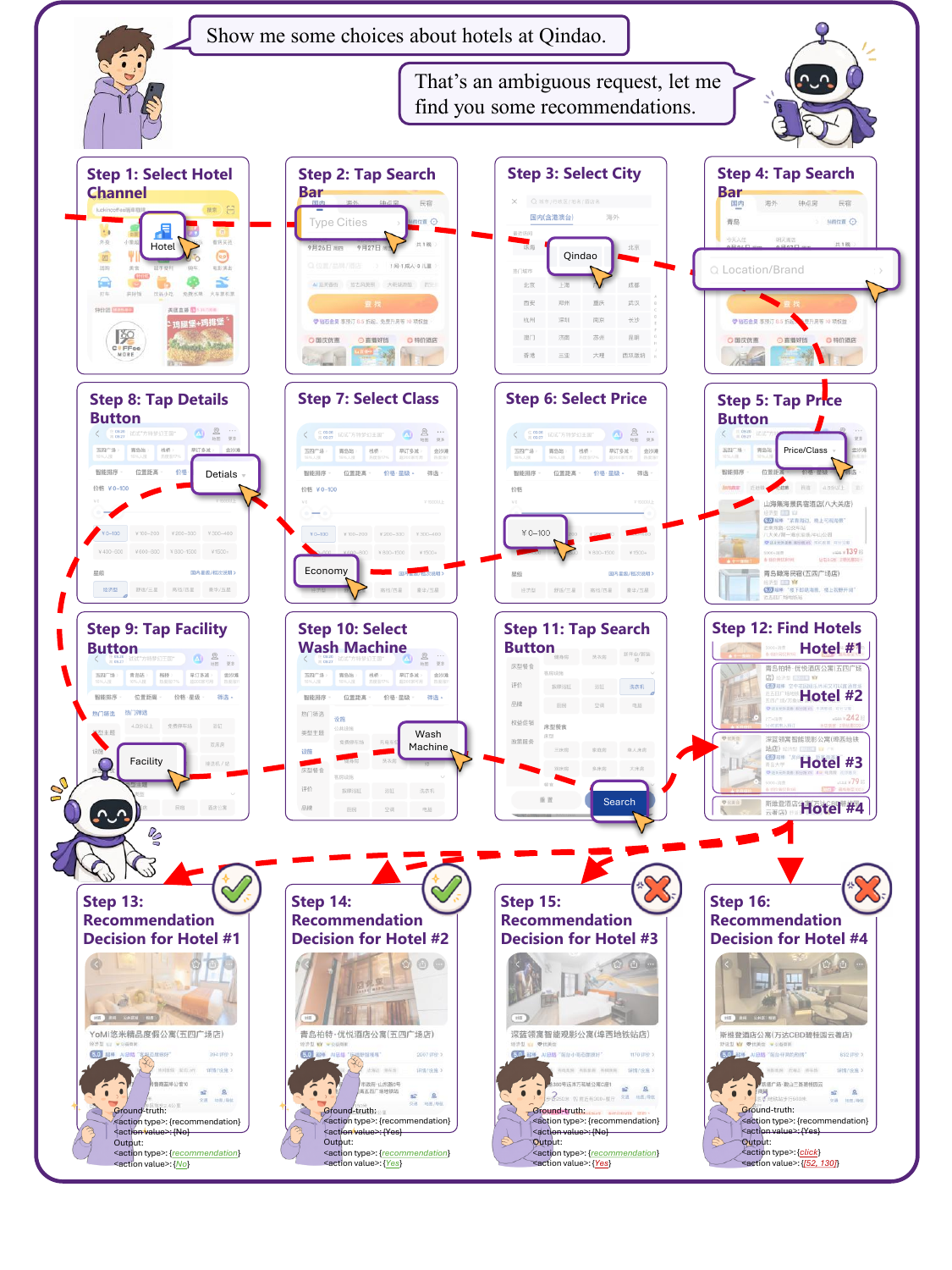}
  \vspace{-5pt}
  \caption{Case Study of our proposed ODYSSE.}
  \label{fig:case_study}
\end{figure}

\clearpage
\twocolumn

\end{document}